\documentclass{article}

\usepackage{PRIMEarxiv}

\usepackage[utf8]{inputenc} 
\usepackage[T1]{fontenc}    
\usepackage{hyperref}       
\usepackage{url}            
\usepackage{booktabs}       
\usepackage{amsfonts}       
\usepackage{nicefrac}       
\usepackage{microtype}      
\usepackage{lipsum}
\usepackage{fancyhdr}       
\usepackage{graphicx}       
\graphicspath{{media/}}     
\usepackage{amsmath}
\usepackage{array}
\usepackage{tabularx}
\usepackage{multirow}
\usepackage{siunitx}
\title{Opening the Black Box of Neural Auction Mechanisms}

\author{
  Li Kong \quad
  Zilong Liu \quad
  Haoyang Zhou \quad
  Hanbing Liu \quad
  Qi Qi\thanks{Corresponding author.} \\
  Gaoling School of Artificial Intelligence \\
  Renmin University of China \\
  Beijing, China \\
  \texttt{kongli@ruc.edu.cn}
  \quad
  \texttt{qi.qi@ruc.edu.cn}
}

\begin{document}
\maketitle

\begin{abstract}
Neural auction design has emerged as a powerful paradigm within automated mechanism design for modern Web platforms. Because these learned mechanisms directly dictate critical economic outcomes, their lack of internal transparency raises significant auditability and trust concerns. Opening this black box \textit{post-hoc} is therefore essential, yet presents two distinctive challenges: (1) activation contexts are dense, correlated numerical profiles devoid of explicit semantic structure, and (2) feature interpretations can only be falsifiably audited if they map to measurable auction states or outcomes.
In this paper, we propose a post-hoc white-box auditing framework for unpacking trained neural auctions. Using CITransNet as a representative testbed, we first employ sparse autoencoders to decompose hidden representations while faithfully preserving downstream auction behavior. Anchoring feature interpretations to native economic targets, we validate them through activation prediction and causal interventions to isolate verified features from the reconstruction dictionary. Finally, applying this pipeline across varied auction settings, we show that the verified features can serve as analytical instruments to examine how internal representations modulate strategic misreporting incentives, align with Myerson's optimal auction, and support the diagnosis of individual auctions. Our work highlights the potential of white-box mechanism auditing to build the foundational trust necessary for neural auctions to be deployed as widely adopted Web technologies. 
To enhance research transparency and facilitate independent inspection, we publicly release the experimental results supporting our findings at \url{https://github.com/Li-Beverly-Kong/opening-neural-auctions.git}.

\end{abstract}


\section{Introduction}

Auctions serve as the foundational economic engine of the modern Web, powering sponsored search and e-commerce marketplaces~\cite{edelman2007internet,varian2007position}. In these platforms, allocation and payment rules directly dictate critical economic outcomes, including advertiser incentives, user welfare, and platform revenue. While classical mechanisms make these relationships explicit~\cite{vickrey1961counterspeculation,myerson1981optimal}, optimal design remains analytically intractable beyond structured settings. Automated mechanism design (AMD) addresses this bottleneck through computational search~\cite{conitzer2002complexity,sandholm2003automated}, with neural auction design emerging as a prominent approach~\cite{curry2025automated}. By parameterizing allocation and payment rules via deep networks, models such as RegretNet and CITransNet~\cite{dutting2019optimal,duan2022context} achieve strong revenue--regret tradeoffs in multi-item and contextual environments. However, despite the expressive power and empirical success of neural auctions, they operate as black boxes, raising critical trust and auditability concerns and hindering their adoption in high-stakes Web transactions. When an algorithm autonomously establishes market rules, opening its black box is no longer a mere technical preference, but an absolute economic necessity.

This necessity shifts the core challenge from maximizing raw performance to auditing how that performance is achieved. To date, efforts to instill trust in neural auctions have largely pursued design-time structure or formal verification. Approaches such as architectural constraints, menu representations, monotonicity enforcement, and certification procedures aim to either shape the mechanism \textit{a priori} or bound the properties of its outputs~\cite{curry2020certifying,peri2021preferencenet,duan2023scalable}. However, hard architectural priors often compromise expressive capacity and revenue, while formal verification struggles to scale. In contrast, we ask a fundamentally different question: how can we audit \textit{what already exists}? Operating strictly post-hoc, our framework unpacks trained, end-to-end neural auctions without altering their original architectures or optimization dynamics.

Because neural auctions compress coupled valuation, allocation, and pricing signals into polysemantic hidden states, sparse autoencoders (SAEs) provide a principled unsupervised approach to disentangle them into interpretable sparse features, as demonstrated in language and biological models~\cite{bricken2023monosemanticity,templeton2026scaling, simon2025interplm}. However, their application to neural auctions presents two distinctive challenges. First, an activation context is a dense and correlated numerical auction profile without explicit semantic structure, making feature semantics difficult to infer directly from top-activating examples. Second, feature interpretations can only be falsifiably audited if they map to measurable auction states or outcomes. To address these challenges, we anchor candidate features directly to the mechanism's native economic targets rather than relying on open-ended discovery. 
Moreover, as unsupervised SAEs capture general input patterns alongside outcome-relevant features~\cite{bricken2023monosemanticity,gao2025scaling,simon2025interplm}, we evaluate each feature along two complementary axes: testing whether its generated explanation reliably predicts held-out activations, and whether scaling its activation causally shifts its designated economic target.

To validate this framework, we focus on Transformer-based neural mechanisms, a prominent architectural paradigm for contextual auction design capable of dynamically integrating bidder, item, and market contexts~\cite{duan2022context,rahme2021permutation}.
We specifically instantiate our audit on CITransNet~\cite{duan2022context}, the canonical, open-source Transformer architecture for contextual auctions. Grounding our audit in fundamental economic dimensions, including incentive compatibility, surplus allocation, and mechanism efficiency, we evaluate our framework across varied auction scales and context domains. Through this approach, we show how a trained neural auction can be turned from a black box into an auditable economic mechanism: one whose sparse features can be grounded in economic targets, interventionally tested, and used to diagnose incentive behavior. By establishing a rigorous foundation for post-hoc white-box auditing, our framework paves the way for deploying neural mechanisms as transparent, accountable, and trustworthy economic technologies in Web platforms.

Our contributions are threefold:
\begin{itemize}
    \item \textbf{White-box auditing for neural auction design.} 
    We pioneer a post-hoc white-box auditing framework for learned mechanisms. Operating strictly post-hoc without altering original architectures or optimization dynamics, our approach provides a methodological foundation for inspecting and intervening on the internal representations of trained neural mechanisms.

    \item \textbf{Causal economic validation.} 
    We establish an evidentiary validation pipeline that anchors feature interpretations to the mechanism's native economic targets. After establishing downstream behavioral faithfulness under SAE reconstruction, we evaluate whether each explanation reliably predicts feature activations and whether intervening on that feature shifts its designated economic target. This protocol filters the reconstruction dictionary to separate features with interventional influence from merely correlational ones, producing verified interpretable representations and economic insights.


    \item \textbf{Actionable mechanism diagnosis.} 
    We demonstrate that the validated interpretable information can serve as analytical instruments for probing the internal economic structure of learned mechanisms. We leverage these features to audit how internal representations affect strategic misreporting incentives, align with canonical optimal-auction theory, and localize realized allocations and payments in individual auctions. This highlights their potential for diagnosing internal bottlenecks in neural mechanisms.

\end{itemize}
\section{Related Work}

\subsection{Neural Auction Design}

Neural auction design formulates mechanism design as a differentiable optimization problem, mapping reported valuations and public context to allocations and payments. Foundational models such as RegretNet~\cite{dutting2019optimal} maximize expected revenue subject to empirical regret penalties, while subsequent architectures~\cite{rahme2021permutation,curry2022learning,duan2022context} incorporate permutation symmetry, matching feasibility, and bidder--item context to broaden the environments in which such mechanisms can be learned. This line of work establishes the performance and expressiveness of neural auction design, but generally evaluates the learned rule through economic quantities computed at its outputs.

To establish trust beyond standard empirical metrics, the community pursues interpretability via \textit{a priori} architectural constraints or formal verification~\cite{curry2020certifying,kuo2020proportionnet,peri2021preferencenet,weissteiner2021monotone,curry2022differentiable,duan2023scalable}. Recent work extends menu-based dominant-strategy incentive compatibility design to large combinatorial outcome spaces~\cite{wang2026bundleflow}, tests distributional assumptions~\cite{kher2025safely}, and certifies revenue upper bounds via neural duality~\cite{jiang2026duality}. Rather than replacing structural guarantees or formal certification, our representation-level audit complements these approaches: operating on a fixed, trained neural auction without requiring architectural constraints or retraining, we examine how internal features represent economic states and how intervening on them affects allocations, payments, and strategic incentives.

\subsection{Post-Hoc Interpretability and SAEs}

Post-hoc interpretability spans several paradigms with distinct analytical scopes. While input attribution~\cite{sundararajan2017axiomatic,lundberg2017unified}, supervised probing~\cite{alain2016understanding,kim2018interpretability}, and dense projections such as PCA either lack internal interventional features, require predefined concept labels, or suffer from polysemantic superposition~\cite{elhage2022toy}, sparse autoencoders learn a sparse basis for hidden activations without supervision, providing candidate units for inspecting representations~\cite{makhzani2013k,bricken2023monosemanticity,gao2025scaling,templeton2026scaling}, exposing domain-relevant features in language and biological models~\cite{simon2025interplm}. These features can be summarized through automated explanation~\cite{bills2023language,paulo2025automaticallyinterpretingmillionsfeatures}.


To elevate these descriptive summaries into reliable explanations, the research has shifted its focus from observational correlations to rigorous causal evaluations, testing whether sparse features generate testable predictions and perform consistently across benchmarks~\cite{paulo2025automaticallyinterpretingmillionsfeatures,movva2025sparseautoencodershypothesisgeneration,kantamneni2025sparseautoencodersusefulcase,karvonen2025saebenchcomprehensivebenchmarksparse}. 
However, standard open-ended semantic discovery remains inadequate for automated mechanism design, where concepts must strictly map to precise economic properties rather than fluid natural-language topics.
To adapt these interpretability foundations for learned auctions, our key departure is to anchor feature interpretations to the mechanism's native economic targets. This bridges representation analysis and economic evaluation, enabling predictive and interventional tests that turn sparse features into practical instruments for mechanism diagnosis.
\section{Methodology}
\label{sec:method}
To audit black-box auction mechanisms, we introduce an interpretability pipeline that links sparse neural representations to falsifiable feature descriptions and target-aligned interventional responses. We first establish a faithful sparse decomposition of internal states, validate the predictive fidelity of their generated explanations, and verify monotonic effects on mechanism outputs under feature interventions~\cite{gao2025scaling}. Finally, we use these features to examine the neural auction mechanism's incentive properties and alignment with optimal-auction theory, and to diagnose observed behavior in individual auctions.

\subsection{Audit Setting and Mechanism Outcomes}
\label{subsec:v2-method-setting}
We consider a contextual auction with $n$ bidders and $m$ items. An auction
instance is represented as
\begin{equation}
    x = (\mathbf{v},\mathbf{b},\mathbf{c}),
    \qquad
    \mathbf{v}\in[0,1]^{n\times m},
    \label{eq:auction-instance}
\end{equation}
where $v_{ij}$ is bidder $i$'s valuation for item $j$, and
$\mathbf{b}$ and $\mathbf{c}$ denote bidder- and item-side contextual
features, respectively. We audit a fixed CITransNet checkpoint
\cite{duan2022context}, which parameterizes the allocation and payment rules as $ M^{\theta}(x) = \bigl(A^{\theta}(x)\in[0,1]^{n\times m},p^{\theta}(x)\in\mathbb{R}_{+}^{n}\bigr).$
Here, $A_{ij}^{\theta}(x)$ is the probability or fractional extent to which
item $j$ is allocated to bidder $i$, and $p_i^{\theta}(x)$ is the payment
charged to bidder $i$. The model parameters $\theta$ remain fixed throughout
the audit.

Under truthful bidding, bidder $i$'s quasilinear utility is given by
\begin{equation}
u_i^\theta(x)=\sum_{j=1}^m v_{ij}A^\theta_{ij}(x)-p_i^\theta(x).
\label{eq:v2-utility}
\end{equation}
We evaluate incentive-related behavior through ex post regret. For a unilateral
misreport $\widehat{\mathbf{v}}_i\in[0,1]^m$ by bidder $i$, define the bidder's realized utility as $u_i^{\theta} \bigl( \mathbf{v}_i; \widehat{\mathbf{v}}_i, \mathbf{v}_{-i}, \mathbf{b}, \mathbf{c} \bigr) = \sum_{j=1}^{m} v_{ij} A_{ij}^{\theta}\bigl(\widehat{\mathbf{v}}_i, \mathbf{v}_{-i}, \mathbf{b}, \mathbf{c} \bigr) - p_i^{\theta} \bigl(\widehat{\mathbf{v}}_i, \mathbf{v}_{-i}, \mathbf{b}, \mathbf{c} \bigr).$
We define ex post regret as the bidder-average maximum utility gain
from unilateral deviations:
\begin{equation}
 \operatorname{Reg}_\theta(x) =\frac1n\sum_{i=1}^n \sup_{\widehat{\mathbf v}_i\in[0,1]^m}  \bigl[u_i^\theta( \mathbf{v}_i; \widehat{\mathbf{v}}_i, \mathbf{v}_{-i}, \mathbf{b}, \mathbf{c} \bigr) -u_i^\theta(x)\bigr].
\label{eq:v2-individual-regret}
\end{equation}
In practice, we approximate the inner supremum via numerical search. Because the candidate deviations include truthful reporting, the resulting empirical regret satisfies $0\leq\widehat{\operatorname{Reg}}_\theta(x) \leq\operatorname{Reg}_\theta(x)$.

To ground feature interpretations in the mechanism's behavior, we define seven economic targets for each auction instance $x$. Let $V_i^\theta(x) = \sum_{j=1}^m v_{ij} A_{ij}^\theta(x)$ denote bidder $i$'s allocated valuation, and $A_{n+1,j}^\theta(x) = 1 - \sum_{i=1}^n A_{ij}^\theta(x)$ be the unallocated probability for item $j$. Let $A_{(1)j}^\theta(x) \geq A_{(2)j}^\theta(x)$ be the two highest allocation probabilities among actual bidders on item $j$. The targets are defined as:
\begin{equation}
\begin{aligned}
 R(x) &= \sum_{i=1}^n p_i^\theta(x), \\
 W(x) &= \sum_{i=1}^n V_i^\theta(x), \\
 U_i(x) &= V_i^\theta(x) - p_i^\theta(x), \\
 P_i(x) &= \frac{p_i^\theta(x)}{V_i^\theta(x)} \quad \text{for } V_i^\theta(x) > 0, \\
 H(x) &= -\frac1m \sum_{j=1}^m \sum_{i=1}^{n+1} A_{ij}^\theta(x) \ln A_{ij}^\theta(x), \\
 L(x) &= \frac1m \sum_{j=1}^m \bigl(A_{(1)j}^\theta(x) - A_{(2)j}^\theta(x)\bigr), \\
 I(x) &= -\widehat{\operatorname{Reg}}_\theta(x).
\end{aligned}
\label{eq:v2-economic-targets}
\end{equation}
Here, the first six targets are direct forward-pass outcomes: revenue $R$ and social welfare $W$ capture aggregate monetization and efficiency; bidder utility $U_i$ and payment share $P_i$ reflect surplus division; and allocation entropy $H$ (with $0\ln 0=0$) and winner margin $L$ quantify allocation uncertainty and competitive separation. By contrast, incentive score $I$ is a counterfactual metric that penalizes searched ex post regret across unilateral deviations.

\subsection{Sparse Decomposition and Evaluation}
\label{subsec:v2-method-sae}
For the fixed mechanism, let $t=(x,i,j)$ index a bidder--item pair and
$h_t\in\mathbb R^d$ denote its hidden representation at the audited layer.
We fit an SAE directly to the bidder--item representations $h_t$. Before encoding, each dimension is standardized using training-set moments, $\widetilde{h}_t=(h_t-\mu)\oslash\sigma$.

The encoder $\operatorname{Enc}:\mathbb{R}^{d}\rightarrow\mathbb{R}^{K}$ maps $\widetilde{h}_t$ to a feature vector $z_t$ by retaining at most the $s$ largest positive activations:
\begin{equation}
    z_t = \operatorname{TopK}_s\bigl(\operatorname{ReLU}(\operatorname{Enc}(\widetilde{h}_t))\bigr), \qquad \|z_t\|_0 \leq s.
\label{eq:sae-encoder}
\end{equation}
The decoder $\operatorname{Dec}:\mathbb{R}^{K}\rightarrow\mathbb{R}^{d}$ reconstructs the standardized representation as $\widehat{\widetilde{h}}_t=\operatorname{Dec}(z_t)$. The primary reconstruction objective is
\begin{equation}
    \mathcal{L}_{\mathrm{rec}}
    =
    \frac{1}{d}
    \mathbb{E}_{t}
    \left\|
        \widetilde{h}_t-\widehat{\widetilde{h}}_t
    \right\|_{2}^{2}.
    \label{eq:sae-reconstruction-loss}
\end{equation}

We use a TopK SAE because it makes the sparsity level an explicit architectural choice~\cite{makhzani2013k,gao2025scaling}, rather than controlling sparsity indirectly through a regularization coefficient.
We set $d=64$, $K=8d=512$, and $s=32$. Although $z_t\in\mathbb{R}^{512}$, at most 32 features are active for any bidder--item pair, while different pairs may activate different subsets of the 512-feature dictionary.

A decomposition used for internal auditing should preserve both the hidden
representations and the downstream behavior of the learned auction. We assess representation reconstruction using the reconstruction loss 
$\mathcal{L}_{\mathrm{rec}}$ from
Eq.~\eqref{eq:sae-reconstruction-loss}, together with
\begin{equation}
    \overline{L}_{0}
    =
    \mathbb{E}_{t}
    \left[
        \lVert z_t\rVert_{0}
    \right],
    f_{\mathrm{inactive}}
    =
    1-
    \frac{1}{K}
    \left|
        \bigcup_{t}
        \operatorname{supp}(z_t)
    \right|,
    \label{eq:representation-faithfulness}
\end{equation}
where $\operatorname{supp}(z_t)=\{k\in[K]:z_{t,k}>0\}$.
Here, $\overline{L}_{0}$ measures the realized number of active features per
bidder--item pair, while $f_{\mathrm{inactive}}$ is the fraction of
dictionary features that do not activate on any evaluated pair.

Representational fidelity alone does not guarantee preservation of the
auction computation. We therefore map each reconstruction back to the original representation scale, $h_t^{\mathrm{SAE}}=\mu+\sigma\odot\widehat{\widetilde{h}}_t$,
and replace the audited hidden states of the frozen CITransNet with
these reconstructions. Let
$M^{\theta}_{\mathrm{SAE}}(x)
=(A^{\mathrm{SAE}}(x),p^{\mathrm{SAE}}(x))$
denote the resulting mechanism. We assess downstream fidelity using
the mean allocation and payment $L_1$ differences:
\begin{equation}
\begin{aligned}
 \text{Allocation }L_1
 &= \mathbb{E}_x\left[
      \sum_{i=1}^{n}\sum_{j=1}^{m}
      \left|A_{ij}^{\mathrm{SAE}}(x)-A_{ij}^{\theta}(x)\right|
    \right],\\
 \text{Payment }L_1
 &= \mathbb{E}_x\left[
      \sum_{i=1}^{n}
      \left|p_i^{\mathrm{SAE}}(x)-p_i^{\theta}(x)\right|
    \right].
\end{aligned}
\label{eq:sae-patch-effect}
\end{equation}

\subsection{Feature Explanations and Predictive Fidelity}
\label{subsec:v2-method-explanations}
Having established the faithfulness of the sparse decomposition, we ask whether its features capture economically meaningful concepts and whether such interpretations capture \textit{when} each feature activates.
For each auction instance $x$ and feature $f \in [K]$, let $z_{ij,f}(x)$ denote the activation of feature $f$ at the bidder--item pair $(i,j)$, and define the auction-level activation as $\bar{z}_f(x) = \max_{i,j} z_{ij,f}(x)$. An auction is active for feature $f$ if $\bar{z}_f(x) > 0$. In active auctions, we identify the focal bidder $i_f^*(x) = \arg\max_i (\max_j z_{ij,f}(x))$ associated with the maximally activating representation. For bidder-level economic targets, we then evaluate the focal utility $U_f(x) = U_{i_f^*(x)}(x)$ and focal payment share $P_f(x) = P_{i_f^*(x)}(x)$ at this bidder.

Before generating explanations, we characterize each feature's economic associations by computing the Spearman rank correlation $\rho_f(Y)$ between auction-level activation $\bar z_f(x)$ and each target $Y(x)$ in Eq.~\eqref{eq:v2-economic-targets}. We compute these coefficients separately on disjoint generation and prediction auction partitions. Auction-level targets are evaluated all auctions, while bidder-level targets are restricted to active auctions and their focal bidders.

We evaluate auction-level targets on all auctions and bidder-level targets only on active auctions, using their focal bidders.

Following language-model explanation and activation-prediction
approaches~\cite{bills2023language,gao2025scaling},
an explainer generates a description $e_f$ and selects an associated
economic target $Y_f$ based on the generation auctions, while a blinded predictor predicts activation on disjoint prediction auctions. We assign $d_f=\operatorname{sign}(\rho_f)$, where
$\rho_f$ is the generation-partition Spearman correlation between
$\bar z_f(x)$ and $Y_f(x)$, whenever this coefficient is finite and
nonzero. The explanation, target, and direction are fixed before
prediction and interventions. For prediction examples $\mathcal S_f$, let
$y_{sf}=\mathbf1[\bar z_f(x_s)>0]$ and
$\widehat y_{sf}\in\{0,1\}$ denote actual and predicted activation.
Pooled precision and recall are
\begin{equation}
 \mathrm{Precision} =\frac{\sum_f\sum_{s\in\mathcal S_f}y_{sf}\widehat y_{sf}}
         {\sum_f\sum_{s\in\mathcal S_f}\widehat y_{sf}}, 
  \mathrm{Recall} =\frac{\sum_f\sum_{s\in\mathcal S_f}y_{sf}\widehat y_{sf}}
         {\sum_f\sum_{s\in\mathcal S_f}y_{sf}}.
\label{eq:v2-explanation-scores}
\end{equation}

\subsection{Causal Interventions and Effects}
\label{subsec:v2-method-responses}
Because activation prediction only characterizes \textit{when} a feature fires, we separately test whether scaling $f$ causally steers its associated target $Y_f$. Specifically, we scale feature $f$'s activation by a dose factor $\gamma \geq 0$ across bidder--item pairs $t=(x,i,j)$:
\begin{equation}
 z_t^{(f,\gamma)} = (z_{t,1}, \ldots, \gamma z_{t,f}, \ldots, z_{t,K}), 
 h_t^{(f,\gamma)} = \mu + \sigma \odot \operatorname{Dec}\bigl(z_t^{(f,\gamma)}\bigr).
\label{eq:v2-coordinate-edit}
\end{equation}
Replacing the audited layer states with $h_t^{(f,\gamma)}$ induces an edited mechanism $M_\theta^{f,\gamma}$, where $\gamma=0$ ablates the feature and $\gamma=1$ recovers the SAE baseline.

With $(e_f, Y_f, d_f)$ and predictive results fixed, we evaluate causal responses on an independent test split, restricting to active auctions $\mathcal{A}_f = \{x : \bar{z}_f(x) > 0\}$ and keeping the focal bidder $i_f^*(x)$ fixed.
The paired causal contrast is
\begin{equation}
 \delta_f(\gamma) = \mathbb{E}\bigl[ Y_f^\gamma(x) - Y_f^1(x) \mid x \in \mathcal{A}_f \bigr],
\label{eq:v2-active-response}
\end{equation}
where $Y_f^\gamma(x)$ evaluates target $Y_f$ under $M_\theta^{f,\gamma}$. Following \cite{gao2025scaling}, comparing against the $\gamma=1$ reconstruction baseline isolates the effect of scaling feature $f$ from the SAE reconstruction residual. 

To quantify the dose-response relationship, we evaluate the mechanism over the ordered dose grid $\Gamma = \{\gamma_1, \gamma_2, \gamma_3, \gamma_4\} = \{0, 0.5, 1, 2\}$. For each auction $x$, the dose-response slope is measured by the OLS slope:
\begin{equation}
 \beta_{xf} = \frac{\operatorname{Cov}_{\gamma \in \Gamma}\bigl(\gamma, Y_f^\gamma(x)\bigr)}{\operatorname{Var}_{\gamma \in \Gamma}(\gamma)}.
\label{eq:v2-response-slope}
\end{equation}
To determine whether feature $f$ causally steers target $Y_f$ in the hypothesized direction $d_f$, we test the directional hypothesis on active auctions:
\begin{equation*}
 H_{0,f}: \mathbb{E}\bigl[d_f \beta_{xf} \mid x \in \mathcal{A}_f\bigr] \leq 0 \quad \text{vs.} \quad H_{1,f}: \mathbb{E}\bigl[d_f \beta_{xf} \mid x \in \mathcal{A}_f\bigr] > 0
\end{equation*}
via a one-sided $t$-test. Let $q_f$ denote the Benjamini--Hochberg (BH)-adjusted $p$-value~\cite{benjamini1995controlling} across all features. We define the filtered strong-response feature set as:
\begin{equation}
\begin{aligned}
\mathcal{S}_{\mathrm{strong}} = \bigl\{f \in [K] : \;& q_f \leq 0.05, \\
& d_f \bigl[\bar{Y}_f(\gamma_{\ell+1}) - \bar{Y}_f(\gamma_\ell)\bigr] > 0, \; \forall \ell \in \{1,2,3\}\bigr\},
\end{aligned}
\label{eq:v2-strong-response}
\end{equation}
where $\bar{Y}_f(\gamma) = \mathbb{E}_{x \in \mathcal{A}_f}[Y_f^\gamma(x)]$ is the mean target at dose $\gamma$.

\subsection{Applications to Mechanism Analysis}
\label{subsec:v2-method-uses}
Beyond validating interpretability, a faithful and causally active feature dictionary provides white-box analytical capabilities for probing a trained neural mechanism. Rather than treating learned rules as opaque artifacts, we exploit these grounded features as analytical instruments across three complementary dimensions: auditing strategic incentive shifts under counterfactual interventions (A1), evaluating alignment with optimal-auction theory (A2), and diagnosing allocation and surplus behaviors in auction cases (A3).

\paragraph{A1: Incentive consequences of outcome-related interventions.}
Because all features in $\mathcal{S}_{\mathrm{strong}}$ are selected via the six direct forward-pass outcomes rather than empirical regret, we next assess how feature scaling affects bidders' incentives for unilateral misreporting. At each dose $\gamma \in \Gamma$, we reoptimize each bidder's report under $M_\theta^{f,\gamma}$ while holding public contexts fixed and competing bids truthful, yielding the counterfactual empirical regret $\widehat{\operatorname{Reg}}_\theta^{f,\gamma}(x)$. Regressing these regret values on dosage $\gamma$ yields an auction-level regret slope $\beta_{xf}^{\mathrm{reg}}$ across active instances. We then conduct two-sided $t$-tests of the null hypothesis of zero mean regret slope with BH adjustment across $\mathcal{S}_{\mathrm{strong}}$, identifying features whose amplification significantly increases or decreases searched deviation gains.

\paragraph{A2: Alignment with optimal-auction theory.}
To test whether the mechanism’s internal representations align with canonical auction structure, we benchmark them against Myerson's revenue-optimal auction~\cite{myerson1981optimal} in a two-bidder, single-item setting with conditionally independent uniform valuations $v_i \sim U[0, \bar{v}_i]$. Myerson's rule allocates to the bidder with the highest non-negative virtual value $\phi_i(v_i) = 2v_i - \bar{v}_i$ and charges a critical payment with $i^* = \operatorname{argmax}_{k \in \{1,2\}} \phi_k$:
\begin{equation}
 A_i^{\mathrm{oracle}} = \mathbf{1}[i=i^*]\,\mathbf{1}[\phi_i \geq 0],
 p_i^{\mathrm{oracle}} = A_i^{\mathrm{oracle}} \frac{\bar{v}_i + \max(0, \phi_{-i})}{2},
\label{eq:v2-myerson-oracle}
\end{equation}
where $\phi_{-i}$ is the rival's virtual value. We examine theoretical alignment in two stages. First, we test whether individual features can distinguish the three oracle-defined allocation regions: no sale, Bidder 0 winning, and Bidder 1 winning. Treating auctions inside a given region as positive examples and the remaining auctions as negative examples, we use continuous feature activations to classify each region and measure accuracy by the area under the curve (AUC). Second, by ablating or amplifying these aligned features, we test whether they move mechanism outputs toward or away from the Myerson oracle, measured by changes in allocation and payment mean absolute error (MAE), relative to the SAE reconstruction.

\paragraph{A3: Diagnostic case studies in individual auctions.}
Finally, we conduct microscopic audits on representative auction instances characterized by extreme market outcomes, such as high revenue or near-no-trade states. For each case, we isolate the five features with the highest activation across all bidder--item pairs. Combining their validated explanations and causal directions with their spatial activation patterns across bidder--item positions, we localize how internal representations organize realized allocations and payments and diagnose decision bottlenecks in individual auctions.

\section{Empirical Evaluation and Mechanism Audits}
\label{sec:results}
We now present empirical results from auditing CITransNet. We first verify that SAE reconstruction preserves downstream mechanism behavior and that generated explanations reliably predict held-out activations. Multi-dose interventions then isolate a verified subset of 165 strong-response features that exert reproducible, monotonic control over their assigned targets, providing the analytical instruments for three downstream audits: strategic regret sensitivity, Myerson-oracle alignment, and instance-level diagnosis. Our code will be made publicly available upon publication.

\begin{table}[t]
\centering
\small
\caption{Representational and downstream output fidelity of the SAE in $2\times5$c.}
\label{tab:v2-sae-quality}
\begin{tabular}{@{}lr@{}}
\toprule
Empirical Diagnostic & Value \\
\midrule
Mean sparsity \(\overline L_0\) & 31.998 \\
Standardized reconstruction MSE & \(4.44\times10^{-5}\) \\
Inactive feature fraction & 0.00977 \\
Mean allocation \(L_1\) difference & 0.003013 \\
Mean payment \(L_1\) difference & 0.000534 \\
\bottomrule
\end{tabular}
\end{table}

\subsection{Evaluation Setup and Reconstruction}
\label{subsec:v2-evaluation}
\paragraph{Evaluation setup.}
Our primary evaluation centers on a pretrained CITransNet~\cite{duan2022context} for two-bidder, five-item auctions with continuous public contexts (\(2\times5\)c). We audit its representations at the first fully connected layer: positioned after contextual self-attention and prior to the output projection heads, this intermediate bottleneck captures the fully integrated economic context where allocation and payment decisions are formed. The TopK SAE mentioned in~\ref{subsec:v2-method-sae} decomposes its bidder--item representations. To benchmark against single-item optimal-auction theory and test whether our auditing protocol generalizes across auction sizes and context modalities, we independently audit two auxiliary mechanisms: single-item continuous ($2\times1$c) and five-item discrete-context ($2\times5$d).

For the primary mechanism, data from the training distribution are partitioned $80{:}20$ across bidder--item pairs for SAE training and representational evaluation. A separate auction-level partition divides explanation generation and disjoint prediction sets. Following standard automated interpretability protocols that pair LLM explanation generation with isolated, zero-shot activation simulation~\cite{bills2023language,gao2025scaling}, we employ \texttt{gpt-5.6-luna} in separate contexts to generate explanations and predict binary activations on disjoint prediction auctions. Downstream reconstruction fidelity and causal dose interventions are evaluated on an independent test split, with all explanations, target choices, and directional hypotheses frozen beforehand. Details are in Appendix~\ref{app:v2-protocol}.

\paragraph{Reconstruction fidelity.}
Table~\ref{tab:v2-sae-quality} reports decomposition fidelity across both hidden representations and downstream mechanism outputs. At the representation level, the 512-feature SAE achieves near-lossless reconstruction with a standardized MSE of \(4.44\times10^{-5}\). It fully utilizes its sparsity budget of 32, reaching a realized mean sparsity of \(\overline L_0 = 31.998\), while more than 99\% of dictionary features activate at least once. Crucially, patching these reconstructed states back into the network produces negligible distortions in downstream behavior, yielding mean allocation and payment \(L_1\) differences of only 0.0030 and 0.0005, respectively. 
Together, these measurements support using the sparse representation to study the fixed mechanism. Additional diagnostics are reported in Table~\ref{tab:v2-reconstruction}.


\begin{figure}[t]
\centering
\includegraphics[width=0.6\columnwidth]{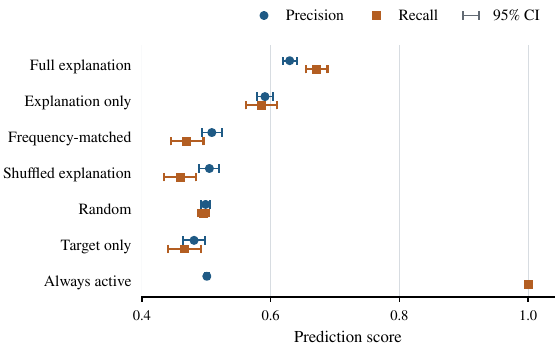}
\caption{Activation-prediction fidelity across all 512 features. Points display pooled precision and recall with 95\% feature-bootstrap confidence intervals.}
\label{fig:v2-explanation-controls}
\end{figure}

\begin{figure}[t]
\centering
\includegraphics[width=\textwidth]{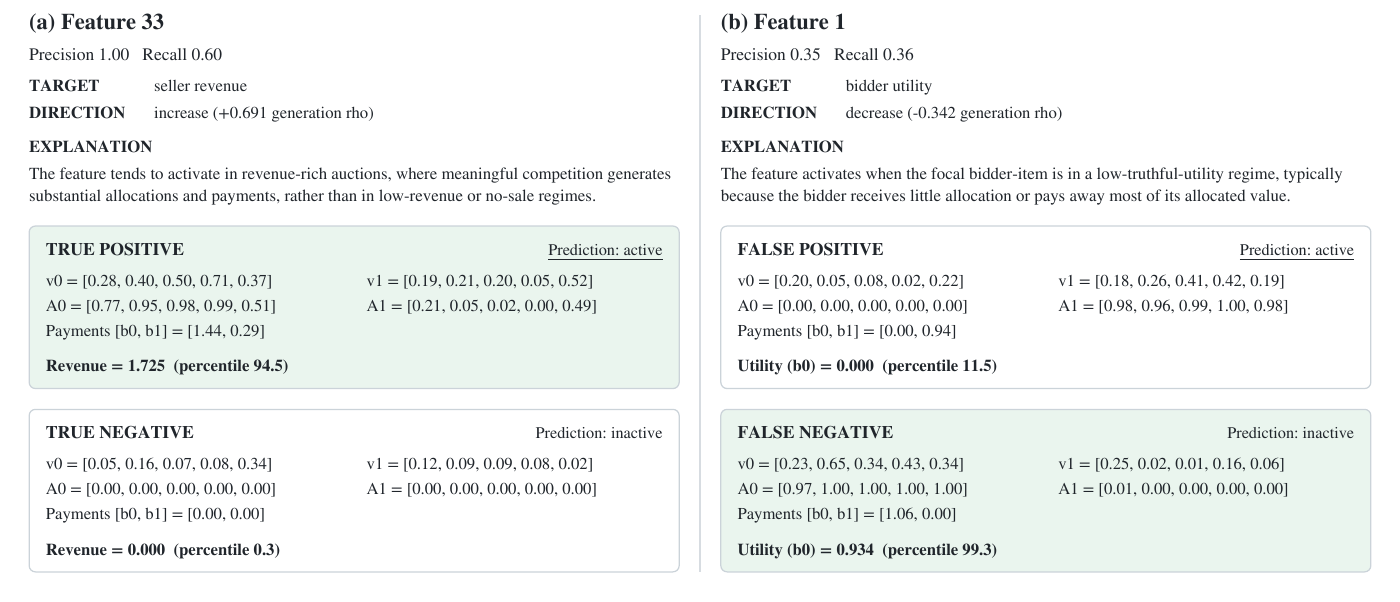}
\caption{Contrasting explanations: F33's predictive revenue description
and F1's overbroad low-utility description. Each column shows the original
text, selected target, preassigned direction, and two auctions in the prediction set with values, allocations, payments, and outcomes. Green marks actual
positive SAE activation, and underlining marks predicted activation.}
\label{fig:v2-explanation-examples}
\end{figure}

\subsection{Predictive Validity of Feature Explanations}
\label{subsec:v2-reliability}
\paragraph{Reproducibility and predictive fidelity.}
Feature--outcome associations remain highly stable across disjoint generation and prediction partitions, with correlations between partition-specific Spearman coefficient vectors ranging from 0.872 to 0.999 across all targets.

Blinded activation predictions across all 512 unfiltered features further evaluate whether the generated explanations capture non-random activation conditions. 
To prevent sparse activation base rates from inflating raw accuracy, we evaluate discriminative validity on a balanced benchmark containing approximately equal numbers of active and inactive auctions for each feature~\cite{paulo2025automaticallyinterpretingmillionsfeatures}.
As illustrated in Figure~\ref{fig:v2-explanation-controls}, even when evaluated across the entire unfiltered dictionary, including features not specialized for downstream economic outputs, the full prompt achieves pooled precision of 0.630 and recall of 0.672 on the balanced benchmark, comparable to dictionary-wide simulation results in language-model SAEs~\cite{gao2025scaling,paulo2025automaticallyinterpretingmillionsfeatures}.

Ablating prompt components demonstrates that the explanation text and target reference provide complementary predictive information.
Relative to the target-only baseline, which removes the text explanation, adding the natural-language description improves precision by 0.149 and recall by 0.206. 
Conversely, relative to the explanation-only setting, which provides text but omits economic target context, incorporating the target reference yields further gains of 0.038 in precision and 0.086 in recall. 
Furthermore, to verify that these predictions reflect semantic comprehension rather than superficial correlation, we compare the full prompt against semantic perturbation controls and trivial baselines.
The full prompt achieves better overall discrimination than shuffled controls that randomly permute explanations across features, frequency-matched controls that substitute text from features with similar activation frequencies, and baseline random or always-active guessing. These consistent advantages demonstrate that the generated explanations capture feature-specific activation conditions, producing reliable and non-random predictions. Complete breakdowns are in Table~\ref{tab:v2-all-controls}.

\paragraph{Contrasting feature explanations.}
Figure~\ref{fig:v2-explanation-examples} illustrates how explanations translate economic associations into concrete activation predictions. For feature F33, which exhibits a Spearman correlation of $+0.691$ with seller revenue on the generation partition, the revenue-rich description cleanly distinguishes a high-revenue active auction from a near-zero-revenue inactive auction. Across the prediction auctions, it identifies 15 of 25 active instances and rejects all 25 inactive instances, achieving a precision of 1.00 and a recall of 0.60. By contrast, F1's low-utility description is flawed because it broadly equates feature activation with low bidder utility, producing 17 false positives and 16 false negatives with a precision of 0.35 and a recall of 0.36. In the displayed auctions, the blinded predictions faithfully follow the description by predicting activation based on the focal bidder's utility, which reverses both ground-truth activation labels. This contrast demonstrates how activation prediction separates an informative feature description from an overbroad restatement of economic correlations.

\begin{figure}[t]
\centering
\includegraphics[width=\textwidth]{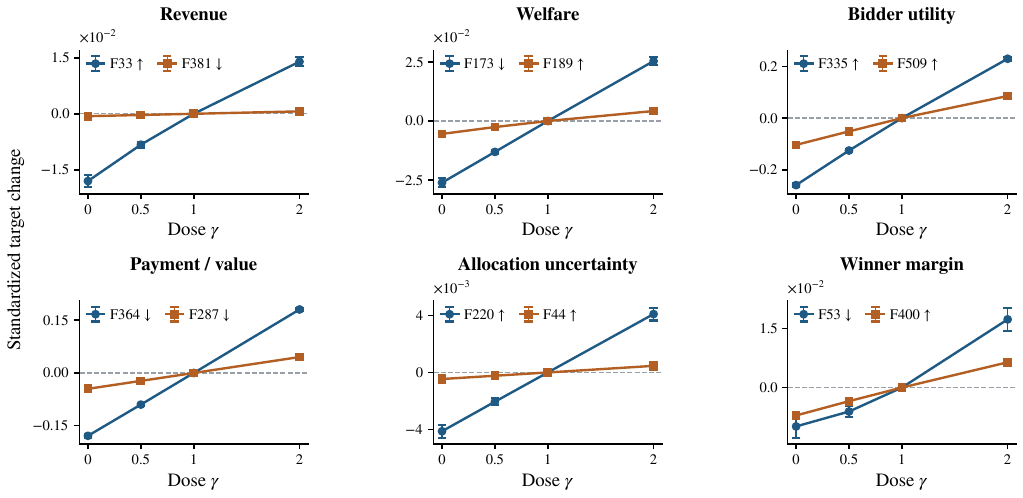}
\caption{Causal dose responses of twelve strong-response features across six economic targets. The ordinate is $d_f\delta_f(\gamma)/\sigma_f$: $\delta_f(\gamma)$ is the mean paired change from SAE reconstruction ($\gamma=1$), $d_f=\pm1$ is the preassigned direction ($\uparrow/\downarrow$), and $\sigma_f$ is the standard deviation of target $Y_f$ across active auctions at the reconstruction baseline. Bars show pointwise 95\% paired $t$ intervals.}
\label{fig:v2-dose-responses}
\end{figure}

\subsection{Causal Responses to Feature Interventions}
\label{subsec:v2-interventions}
Evaluating the learned mechanism under the four-dose intervention protocol reveals that internal representations exert causal control over auction outcomes. Testing all 512 features against the strong-response criteria in Eq.~\eqref{eq:v2-strong-response} isolates \textbf{165 strong-response features}. Consistent with unsupervised SAEs in language and biological models~\cite{bricken2023monosemanticity,gao2025scaling,simon2025interplm}, where many features model input or contextual structure without specializing in specific downstream outputs, recovering 165 features that monotonically steer native economic targets without outcome supervision provides a rich set of verified probes for mechanism analysis. Furthermore, extending the intervention grid to an extreme dosage of $\gamma=8$ retains 158 of these 165 features, shown in Appendix~\ref{app:v2-reconstruction-explanation-results}, confirming that their monotonic responses persist well beyond the primary window.

Figure~\ref{fig:v2-dose-responses} highlights three properties of these causal responses on the held-out test split. First, the curves exhibit clear directional monotonicity: as the intervention dose increases, each target shifts consistently along its preassigned direction $d_f$, establishing a clear feature--outcome correspondence. Second, although our selection criteria only require monotonicity, 163 of the 165 strong-response features exhibit near-linear trajectories where a linear fit of the mean response $\bar{Y}_f(\gamma)$ against dose $\gamma$ achieves $R^2 \geq 0.95$. Rather than causing an abrupt jump only upon ablation, scaling feature strength produces proportional changes in the outcome, making both the direction and the magnitude of the effect directly predictable. Third, the responses translate into direct economic effects. Utility gains indicate more surplus retained by the focal bidder, while payment-share reductions indicate a smaller fraction of allocated value paid to the seller. Features associated with the same target can also have opposite effects, as illustrated by F173 reducing welfare and F189 increasing it.
Notably, empirical regret is not a direct mechanism output of a single forward pass, as it compares truthful utility against reoptimized deviation gains under the edited mechanism. We therefore treat searched regret as a counterfactual check on these interventions and examine how all 165 outcome-related features affect it in Section~\ref{subsec:v2-usefulness}. Representative selection and frozen explanations appear in Appendix~\ref{app:v2-reconstruction-explanation-results}.

Together, faithful reconstruction, activation prediction, and dose interventions link internal representations to interpretable economic responses. The resulting strong-feature subset exhibits statistically supported monotonic effects on preassigned targets, providing a causally grounded basis for subsequent mechanism analysis.


\begin{table}[t]
\centering
\small
\setlength{\tabcolsep}{2.8pt}
\caption{A1: Empirical-regret slope summary across the 165 strong-response features under reoptimized misreports. Raw $\uparrow/\downarrow$ and BH $\uparrow/\downarrow$ count features with positive (increasing) and negative (reducing) mean slopes $\bar{\beta}_f^{\mathrm{reg}}$ in total and at two-sided BH-adjusted $q\leq0.05$. Max $\uparrow$ and Max $\downarrow$ report the extreme slopes per unit dose within each target.}
\label{tab:v2-regret-families}
\begin{tabular}{@{}lrrrrr@{}}
\toprule
Target & $N$ & Raw $\uparrow/\downarrow$ & BH $\uparrow/\downarrow$ & Max $\uparrow$ & Max $\downarrow$ \\
\midrule
Allocation entropy & 6 & 2/4 & 0/1 & $1.27\!\times\!10^{-6}$ & $-3.44\!\times\!10^{-6}$ \\
Bidder utility & 22 & 12/10 & 3/0 & $1.14\!\times\!10^{-2}$ & $-2.49\!\times\!10^{-6}$ \\
Payment share & 89 & 48/41 & \textbf{17/11} & $1.38\!\times\!10^{-2}$ & $-4.30\!\times\!10^{-4}$ \\
Seller revenue & 20 & 12/8 & 3/1 & $1.93\!\times\!10^{-4}$ & $-1.93\!\times\!10^{-5}$ \\
Social welfare & 8 & 6/2 & 3/1 & $5.59\!\times\!10^{-4}$ & $-5.23\!\times\!10^{-4}$ \\
Winner margin & 20 & 8/12 & 0/3 & $2.61\!\times\!10^{-6}$ & $-8.64\!\times\!10^{-4}$ \\
\midrule
Total & 165 & 88/77 & 26/17 & --- & --- \\
\bottomrule
\end{tabular}
\end{table}

\subsection{Applications to Mechanism Analysis}
\label{subsec:v2-usefulness}
Building on these causally grounded features, we now carry out the three mechanism analyses introduced in Section~\ref{subsec:v2-method-uses}.

\paragraph{A1: Incentive consequences of outcome-related interventions.}
Table~\ref{tab:v2-regret-families} examines how steering the 165 strong-response features affects empirical regret under reoptimized misreports. Across this subset, mean empirical-regret slopes are positive for 88 features and negative for 77, with two-sided Benjamini--Hochberg tests at $q \leq 0.05$ identifying 26 regret-increasing and 17 regret-reducing features. Across targets, regret sensitivity is concentrated in surplus and pricing representations: the largest positive slopes for payment share at $1.38\times10^{-2}$ and bidder utility at $1.14\times10^{-2}$ exceed those for revenue, welfare, margin, and entropy by two to four orders of magnitude. While all three significant utility features increase regret, the 89 payment-share features include both 17 regret-increasing and 11 regret-reducing features, showing that pricing-related representations affect strategic incentives in both directions. Notably, the leading surplus-enhancing representatives from Figure~\ref{fig:v2-dose-responses}, utility feature F335 and payment-share feature F364, account for the steepest positive regret slopes in their respective families, with high-dose scaling $\gamma=2$ further compounding mean empirical regret by 0.0331 and 0.0296 above the reconstruction baseline. Applying the same audit to the independently trained $2\times1$c and $2\times5$d mechanisms similarly recovers setting-specific strong-response features with bidirectional regret sensitivity, as reported in Appendix~\ref{app:v2-additional-settings}. Together, these results show how interpretable features support mechanism analysis: interventions on these features identify directions that increase or reduce empirical regret and reveal which internal adjustments strongly amplify or suppress misreporting gains.

\begin{figure}[!t]
\centering
\includegraphics[width=\textwidth]{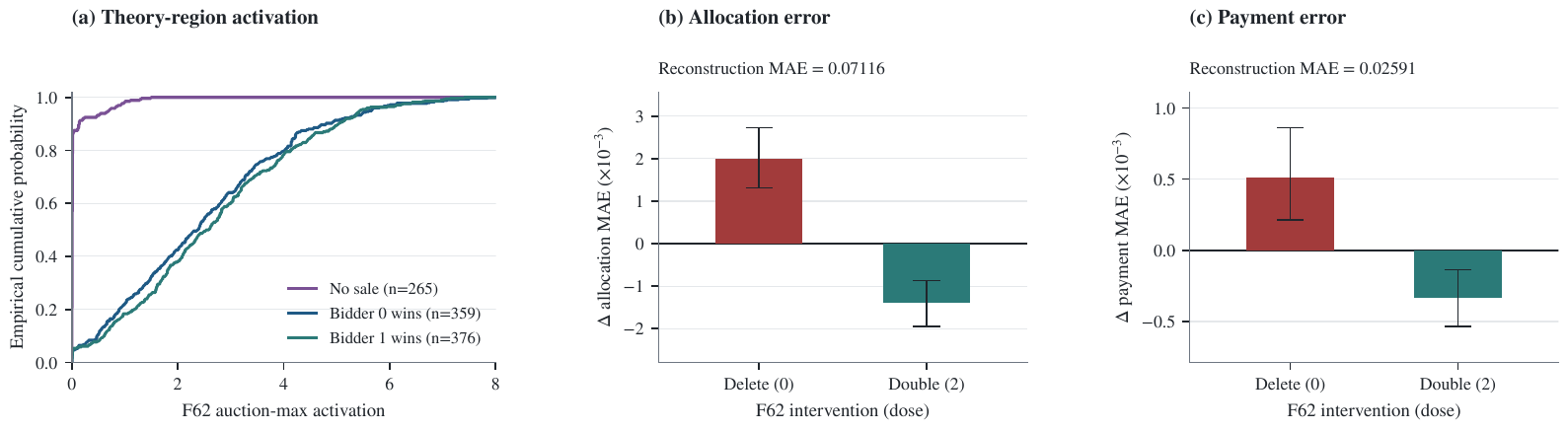}
\caption{Alignment with optimal-auction theory for feature F62 in the $2\times1$c setting. (a) Empirical cumulative distribution of F62 auction-maximum activation across the three Myerson oracle regions on 1,000 evaluation auctions. (b, c) Paired changes in allocation and payment MAE relative to the $\gamma=1$ reconstruction baseline when ablating at dose 0 or doubling at dose 2, with 95\% auction-bootstrap confidence intervals. Negative values indicate closer agreement with the Myerson oracle.}
\label{fig:v2-myerson-interventions}
\end{figure}

\paragraph{A2: Alignment with optimal-auction theory.}
In the single-item $2\times1$c setting, we investigate whether the mechanism's unsupervised representations organically recover canonical economic structure, benchmarking them against the Myerson optimal auction in Eq.~\eqref{eq:v2-myerson-oracle}.
First, individual sparse features accurately identify the prescribed theoretical regions. As shown in Figure~\ref{fig:v2-myerson-interventions}(a), the auction-level activation of feature F62 is zero or near zero in the no-sale region while taking substantially higher values in both bidder winning regions, distinguishing no-sale outcomes with an AUC of 0.9636. This behavior directly matches F62's explanation, which states that the feature \emph{``marks competitively valuable single-item auctions where allocation to a winner generates substantial seller revenue, activating for either bidder's token; low-revenue auctions usually lack it.''} For winner identification, bidder-specific pair readouts of features F73 and F421, both explained as activating when the focal bidder wins an item at a small payment share with substantial surplus, identify the Bidder 0 and Bidder 1 winning regions with AUCs of 0.9129 and 0.9277, respectively. Complete results appear in Appendix Table~\ref{tab:v2-myerson-recovery}.

Second, intervening on this region-aligned representation probes its functional role in approximating the Myerson oracle. Relative to the reconstruction baseline in Figures~\ref{fig:v2-myerson-interventions}(b, c), ablating F62 increases allocation MAE by $1.99\times10^{-3}$ and payment MAE by $0.51\times10^{-3}$, whereas doubling F62 reduces these errors by $1.40\times10^{-3}$ and $0.33\times10^{-3}$. Protocol details are reported in Appendix~\ref{app:v2-myerson}. Interpretability thus grounds abstract theoretical benchmarks in internal features, enabling us to analyze which theoretical states the mechanism represents and which internal adjustments affect alignment with the Myerson oracle.

\paragraph{A3: Diagnostic case studies in individual auctions.}
To illustrate instance-level diagnosis, we inspect two contrasting auctions during unedited forward passes and examine their five most active features across the complete dictionary, with details reported in Appendix~\ref{app:v2-diagnostics}. This audit links which economic features are active, where they activate across bidder--item positions, and how they localize realized allocations and payments.

In the high-revenue auction \texttt{034928}, the top features bring together three causal directions: increasing bidder utility via F509 and F335, decreasing payment share via F419 and F428, and increasing winner margin via F132. Their spatial activation patterns reflect how these roles are organized. Features F509 and F419 activate on each item's winning bidder, covering Bidder 0 on Item 1 and Bidder 1 on Items 0, 2, 3, and 4, whereas F132 and F428 occupy the unallocated side of each item. This configuration couples surplus-related features on the recipients with decisive allocation signals on the unallocated side. By assigning each item decisively to its higher-valuing bidder, the mechanism creates substantial total welfare $W=3.451$, which provides the value base for the seller to collect high revenue $R=2.629$. The diagnostic thus localizes how allocation and surplus representations spatially partition to support high revenue generation.

In the near-no-trade auction \texttt{010715}, the top activations bring together revenue-reducing feature F76, welfare-reducing feature F173, and payment-share-increasing features F283, F16, and F506. Feature F76 activates across all ten bidder--item positions, reflecting an auction-wide no-trade regime, while F283, lying outside $\mathcal{S}_{\mathrm{strong}}$, joins F16 and F506 in marking allocations where payments consume almost all allocated value. These features correspond to severe trade suppression: allocations remain negligible, and revenue and welfare are approximately $2.4\times10^{-4}$. Although the mechanism captures nearly all allocated value as payment, seller revenue remains near zero because transaction scale collapses. This diagnosis highlights that the failure to clear trade, rather than weak extraction, is the primary bottleneck for subsequent mechanism improvement.

These case studies demonstrate how sparse autoencoders enable white-box diagnosis of individual mechanism decisions.
\section{Conclusion}




This paper introduces a post-hoc white-box auditing framework for unpacking trained neural auctions. Through faithful SAE decomposition, blinded activation prediction, and multi-dose interventions, we isolate a substantial set of sparse features from the unsupervised reconstruction basis that exert monotonic, near-linear causal effects over native economic targets. Across our evaluated mechanisms, these verified features serve as practical diagnostic tools: revealing bidirectional effects on strategic regret, aligning with and steering outputs toward Myerson's optimal auction, and pinpointing failure bottlenecks in individual auctions.

Our findings also highlight future directions from post-hoc diagnosis toward mechanism refinement. First, the near-linear dose responses and separation between regret-reducing and regret-increasing directions suggest a basis for constrained feature steering that jointly balances revenue and incentive compatibility. Second, our Myerson audit shows that a neural auction, trained and decomposed without oracle supervision, spontaneously encodes canonical reserve-price and winner boundaries. Beyond settings with known optima, this approach may help generate hypotheses for theoretical mechanism design in analytically intractable multi-item settings. Looking ahead, our work takes a step toward building transparent, steerable, and trustworthy neural auction mechanisms.

\clearpage
\bibliographystyle{unsrt}  
\bibliography{references}  

\clearpage
\appendix
\section{Supplementary Evaluation Setup}
\label{app:v2-protocol}
\subsection{Data Partitions and SAE Training}
Our experimental workflow establishes data isolation across three distinct auditing phases. First, for SAE fitting and representation evaluation, we draw bidder--item representations from the mechanism's training distribution and partition them into an $80{:}20$ split. Second, for feature explanation and predictive verification, a separate auction-level partition divides auctions into approximately equal generation and prediction sets. Third, downstream fidelity and dose interventions are evaluated on held-out test auctions. Each auxiliary setting follows this identical partition protocol with its own mechanism, SAE checkpoint, and feature explanations.

To prevent features from remaining persistently inactive during training, we augment the standard reconstruction objective $\mathcal{L}_{\mathrm{rec}}$ with the auxiliary TopK loss~\cite{gao2025scaling}:
\begin{equation}
 \mathcal{L}_{\mathrm{SAE}} = \mathcal{L}_{\mathrm{rec}}
 + 0.003 \cdot \mathbb{E}_t \left[
 \frac{\|W_{\mathrm{dec}}z_t^{\mathrm{aux}} - \varepsilon_t\|_2^2}
      {\max(\|\varepsilon_t\|_2^2, 10^{-8})} \right].
\label{eq:app-v2-sae-objective}
\end{equation}
Here $W_{\mathrm{dec}}$ denotes the SAE decoder weight matrix, and $\varepsilon_t = \widetilde{h}_t - \widehat{\widetilde{h}}_t$ is the primary reconstruction residual with gradients stopped. To revive dead features, $z_t^{\mathrm{aux}}$ selects up to 128 persistently inactive features based on their pre-activation encoder values and optimizes them to reconstruct this remaining error. This auxiliary branch is \textbf{training-only}: all downstream evaluations and interventions rely on the sparse reconstructions $\widehat{\widetilde{h}}_t$.

\subsection{Explanation Generation and Prediction}
\paragraph{Generation protocol.}
Prompts for explanation generation combine top-activating bidder--item pairs with randomly sampled inactive pairs from the generation partition. For each exemplar, the input context provides the focal bidder--item pair $(i,j)$, its raw activation value $z_{ij,f}(x)$, the complete auction profile, and percentile ranks for all seven economic targets evaluated over the generation distribution. The explainer produces a concise text $e_f$ up to 35 words and selects an associated target $Y_f$. The sign of their generation Spearman correlation determines $d_f$, and the resulting tuple $e_f, Y_f, d_f$ is frozen before evaluation.

\paragraph{Prediction protocol.}
The prediction benchmark balances active auctions using their maximally activating pair against inactive auctions with zero activation throughout. The evaluator receives the generated explanation, the complete auction profile, the focal bidder--item pair $(i,j)$, and the percentile rank of $Y_f$. The output is a binary activation judgment.

\paragraph{Ablation controls and uncertainty.}
To verify that prediction accuracy is reliable and non-random, we benchmark performance by perturbing or removing visible components of the evaluator's input. We evaluate component ablations omitting either the target reference or the explanation text, semantic controls replacing explanations with shuffled or frequency-matched texts from other features, and trivial guessing baselines. To quantify statistical uncertainty, we apply paired feature-level bootstrap resampling to compute 95\% confidence intervals for both absolute metrics and the performance gains of the full prompt over each control.

\subsection{Intervention Evaluation Protocol}
\label{app:v2-incentives}
For bidder-specific targets, the focal bidder $i_f^*(x)$ identified at baseline remains fixed across all doses. Because payment share is undefined when allocated value is zero, paired contrasts and regression slopes strictly evaluate auctions with finite outcomes across all required doses, whereas mean-dose curves average over available instances at each dose. All effect sizes, except for the empirical regret, are standardized by the target standard deviation computed on active auctions under the reconstruction baseline at dose one.

Empirical regret is estimated through multi-start gradient-based optimization over unilateral misreports within the unit cube $[0,1]^m$ using three independent repeats of 20 shared initializations and 200 projected Adam steps with a learning rate of 0.001, holding rival bids and public contexts truthful. To reduce search variance across intervention levels, counterfactual regret evaluations rerun this search under each edited mechanism using identical initial report candidates across all doses.

\section{Supplementary Results}
\subsection{Supplementary Results for Interpretations}
\label{app:v2-reconstruction-explanation-results}
\paragraph{Reconstruction and explanation reliability.}
This part presents the supplementary diagnostics referenced in Sections~\ref{subsec:v2-evaluation} and~\ref{subsec:v2-reliability}. First, complementing the allocation and payment reconstruction metrics in Section~\ref{subsec:v2-evaluation}, Table~\ref{tab:v2-reconstruction} confirms that substituting original hidden states with SAE reconstructions induces negligible shifts across aggregate economic targets, with mean signed differences on the order of $10^{-6}$ to $10^{-5}$. Second, expanding on Figure~\ref{fig:v2-explanation-controls}, Table~\ref{tab:v2-all-controls} reports exact pooled precision and recall across all seven evaluation conditions, showing that the full explanation achieves 95\% confidence intervals of $[0.619, 0.640]$ for precision and $[0.655, 0.688]$ for recall, whose lower bounds strictly exceed the upper bounds of every informative control. This non-overlapping separation confirms that the predictive gains are statistically robust across the dictionary.


\begin{table}[t]
\centering
\small
\caption{Downstream reconstruction differences across auction-level economic targets in the $2\times5$c setting.}
\label{tab:v2-reconstruction}
\begin{tabular}{@{}lr@{}}
\toprule
Economic Target & Mean Difference $\Delta$ \\
\midrule
Revenue $R$ & $-1.57\times10^{-5}$ \\
Welfare $W$ & $-2.46\times10^{-6}$ \\
Allocation entropy $H$ & $-1.27\times10^{-5}$ \\
Winner margin $L$ & $+1.23\times10^{-5}$ \\
\bottomrule
\end{tabular}
\end{table}

\begin{table}[t]
\centering
\small
\caption{Complete breakdown of pooled activation-prediction performance across all 512 features on the balanced prediction benchmark, corresponding to Figure~\ref{fig:v2-explanation-controls}.}
\label{tab:v2-all-controls}
\begin{tabular}{@{}lrrcc@{}}
\toprule
& & & \multicolumn{2}{c}{95\% CI} \\
\cmidrule(l){4-5}
Condition & P & R & P & R \\
\midrule
Full explanation & \textbf{0.630} & 0.672 & [0.619, 0.640] & [0.655, 0.688] \\
Explanation only & 0.591 & 0.586 & [0.579, 0.604] & [0.561, 0.610] \\
Frequency-matched & 0.509 & 0.470 & [0.493, 0.524] & [0.445, 0.496] \\
Shuffled explanation & 0.505 & 0.460 & [0.489, 0.520] & [0.434, 0.484] \\
Random & 0.499 & 0.495 & [0.492, 0.506] & [0.487, 0.504] \\
Target only & 0.481 & 0.466 & [0.464, 0.498] & [0.441, 0.492] \\
Always active & 0.501 & \textbf{1.000} & [0.497, 0.505] & [1.000, 1.000] \\
\bottomrule
\end{tabular}
\end{table}

\begin{table*}[t]
\centering
\footnotesize
\caption{Strong-response representatives and frozen explanations.
$N_a$: active auctions; P/R: explanation precision/recall;
$\beta_{\rm std}$: aligned slope per unit dose, scaled by each feature's
active-auction reconstruction-baseline target SD.}
\label{tab:v2-representatives}
\begin{tabular}{@{}lrccp{0.10\textwidth}p{0.61\textwidth}@{}}
\toprule
Feature & $N_a$ & P/R & $\beta_{\rm std}$ & Target/direction & Frozen explanation \\
\midrule
F33 & 1837 & 1.00/0.60 & 0.01579 & Revenue $\uparrow$ & The feature tends to activate in revenue-rich auctions, where meaningful competition generates substantial allocations and payments, rather than in low-revenue or no-sale regimes. \\
F381 & 389 & 0.74/0.80 & 0.000641 & Revenue $\downarrow$ & The feature activates when low seller revenue reflects weak monetization: competition produces little payment, while allocation and payment mass remain sparse or one-sided around the focal bidder-item. \\
F173 & 1983 & 1.00/0.60 & 0.02578 & Welfare $\downarrow$ & Activates in inefficient allocation regimes where probability mass reaches lower-valued bidders, reducing truthful social welfare; allocations aligned with high valuations more often leave the feature inactive. \\
F189 & 1637 & 0.68/0.92 & 0.004805 & Welfare $\uparrow$ & Activation marks a focal bidder-item token when the focal bidder is clearly outvalued and loses that item; the rival's valuation advantage directs allocation away, typically within relatively high-social-welfare auctions. \\
F335 & 1975 & 1.00/0.60 & 0.24293 & Utility $\uparrow$ & The feature activates when the focal bidder is a near-monopoly winner whose dominant allocation and small payment produce high truthful utility, unlike diffuse or costly allocations. \\
F509 & 1990 & 1.00/0.84 & 0.09413 & Utility $\uparrow$ & Allocating valuable items to the focal bidder while leaving meaningful surplus raises truthful utility; this high-surplus regime predicts activation of the feature. \\
F364 & 1967 & 0.94/0.68 & 0.18009 & Payment/value $\downarrow$ & The feature activates when the focal bidder's payment is small relative to allocated value, whereas paying a large share of value predicts inactivity. \\
F287 & 1927 & 1.00/0.64 & 0.04528 & Payment/value $\downarrow$ & The feature activates when the focal bidder receives valuable allocation while paying only a small fraction of that allocated value, producing high utility rather than the costly allocations typical of inactive tokens. \\
F220 & 1943 & 1.00/0.68 & 0.004116 & Uncertainty $\uparrow$ & The feature activates for focal bidder-item tokens in high-allocation-uncertainty auctions, where diffuse bidder or no-sale probabilities leave the focal item without a decisive winner. \\
F44 & 20 & 0.64/0.72 & 0.000463 & Uncertainty $\uparrow$ & The feature marks focal bidder-item tokens in auctions with diffuse, high-entropy allocations: probability mass is spread across bidders or no-sale rather than a decisive winner, creating an ambiguous allocation regime. \\
F53 & 2000 & 0.88/0.60 & 0.01389 & Margin $\downarrow$ & The feature activates on bidder-item tokens in relatively contested auctions: a smaller winner margin leaves the focal item's allocation less decisive and predicts activation. \\
F400 & 1736 & 0.75/0.60 & 0.006723 & Margin $\uparrow$ & Activation is associated with focal bidder-item tokens in relatively decisive markets, where competition resolves toward a clear winner rather than diffuse sharing, making winner margin the strongest predictor. \\
\bottomrule
\end{tabular}
\end{table*}

\paragraph{Feature interventions.}
\label{app:v2-interventions}
This subsection provides the extended-dose evaluations and representative feature details referenced in Section~\ref{subsec:v2-interventions}. First, when expanding the four-dose intervention grid to the six-dose range $\{0, 0.5, 1, 2, 4, 8\}$, recomputing the directional tests across the dictionary yields 492 finite hypotheses, 162 Benjamini--Hochberg discoveries, and 158 retained strong-response features, corresponding to a 95.8\% retention rate. Only seven features from the primary four-dose set drop out at extreme amplification due to response saturation, directional reversal, or reduced significance, while all downstream analyses retain the primary four-dose selection. Second, to select the twelve representative trajectories displayed in Figure~\ref{fig:v2-dose-responses}, we filter the 165 strong-response features for explanation precision and recall of at least 0.60, which yields 83 features, and choose the two features with the largest aligned standardized slopes within each of the six targets. Table~\ref{tab:v2-representatives} reports the active auction counts, predictive fidelity metrics, aligned standardized slopes, and frozen explanations for these representatives.

\subsection{Alignment with Optimal-Auction Theory}
\label{app:v2-myerson}
This section details the evaluation protocol and complete region-identification results supporting the $2\times1$c Myerson analysis in Section~\ref{subsec:v2-usefulness}. We partition 2,000 test auctions evenly into 1,000 discovery auctions and 1,000 evaluation auctions, where the evaluation split contains 265 no-sale, 359 Bidder-0-win, and 376 Bidder-1-win instances. For each theoretical region and activation level in Table~\ref{tab:v2-myerson-recovery}, we select the feature and high- or low-activation orientation that maximize one-versus-rest AUC on the discovery split and then compute the reported AUC on the evaluation split.

Comparing the two readout levels in Table~\ref{tab:v2-myerson-recovery} reveals how the mechanism organizes global versus bidder-specific theoretical states. At the Auction level, taking the maximum activation $\bar{z}_f(x) = \max_i z_{i1,f}(x)$ across bidders accurately captures the auction-wide no-sale state through revenue feature F62, but yields modest AUCs of 0.6789 and 0.7161 for the two winning regions because taking the maximum across bidders does not distinguish which bidder has the higher activation. By contrast, evaluating activation $z_{i1,f}(x)$ at the Bidder level preserves positional identity, allowing surplus-related payment-share features F73 and F421 to separate each bidder's winning region with AUCs above 0.91. These readouts connect revenue-related features to Myerson's trade/no-trade distinction and bidder-specific features to winner identity, making theoretical allocation states interpretable.

\begin{table}[t]
\centering
\small
\caption{Myerson-region AUCs in the $2\times1$c setting. High and low indicate that feature activation tends to be higher or lower within the region than outside it.}
\label{tab:v2-myerson-recovery}
\begin{tabular}{@{}llllr@{}}
\toprule
Region & Level & Feature & Target & AUC \\
\midrule
No sale & Auction & F62 (low) & Seller revenue & 0.9636 \\
Bidder 0 wins & Auction & F233 (high) & Payment share & 0.6789 \\
Bidder 1 wins & Auction & F62 (high) & Seller revenue & 0.7161 \\
\midrule
Bidder 0 wins & Bidder & F73 (high) & Payment share & 0.9129 \\
Bidder 1 wins & Bidder & F421 (high) & Payment share & 0.9277 \\
\bottomrule
\end{tabular}
\end{table}

\subsection{Diagnostic Cases in Individual Auctions}
\label{app:v2-diagnostics}
\begin{figure}[t]
\centering
\includegraphics[width=\textwidth]{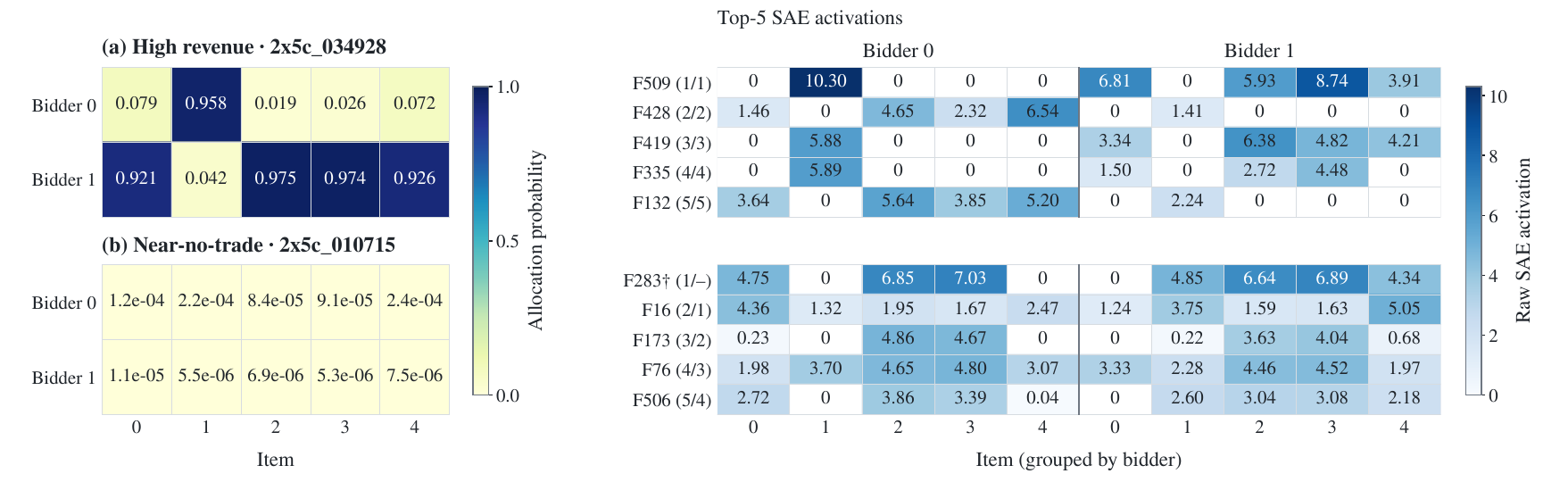}
\caption{Diagnostic case studies in the $2\times5$c setting. \textbf{Left:} Bidder--item allocation probabilities on a shared scale, with scientific notation preserving tiny nonzero allocations. \textbf{Right:} Activation matrices for the top five SAE features in each auction grouped by bidder, where labels ranks as (whole dictionary / strong-response subset) and the dagger $\dagger$ marks F283 outside $\mathcal{S}_{\mathrm{strong}}$.}
\label{fig:v2-case-diagnostics}
\end{figure}

This subsection provides the supporting records for the individual auction diagnostics in Section~\ref{subsec:v2-usefulness}. For each of the two auctions, we run an unedited forward pass and select the five features with the highest auction-level activation $\bar{z}_f(x)$ across all 512 dictionary features, using $\dagger$ to mark F283 because it lies outside $\mathcal{S}_{\mathrm{strong}}$. Figure~\ref{fig:v2-case-diagnostics} compares the bidder--item allocation probabilities with the pair-level activation matrices of these top five features on shared color scales, while Table~\ref{tab:v2-case-details} reports the item valuations, bidder-level accounting, and verbatim explanations. In Table~\ref{tab:v2-case-details}, allocated values $V_i$ and payments $p_i$ are summed across items for each bidder so that $U_i = V_i(1 - P_i)$ for $V_i > 0$, and bidder-level deviation gains are taken from the baseline misreport search in Appendix~\ref{app:v2-incentives}, with empirical regret given by their two-bidder mean.

\begin{table}[t]
\centering
\small
\setlength{\tabcolsep}{4pt}
\caption{Cross-setting causal-response and empirical-regret audits under the four-dose protocol. Strong counts monotonic strong-response features, Regret $\uparrow/\downarrow$ reports significantly increasing/decreasing regret-sensitive features within each strong-response set, and Payment-share $\uparrow/\downarrow$ reports the corresponding counts within the payment-share target.}
\label{tab:v2-setting-audits}
\begin{tabular}{@{}lrrr@{}}
\toprule
Setting & Strong & Regret $\uparrow/\downarrow$ & Payment-share $\uparrow/\downarrow$ \\
\midrule
$2\times5$c & 165 & 26/17 & 17/11 \\
$2\times1$c & 106 & 37/43 & 23/27 \\
$2\times5$d & 142 & 43/10 & 19/3 \\
\bottomrule
\end{tabular}
\end{table}

\begin{table}[t]
\centering
\footnotesize
\caption{Case-level valuations and bidder accounting, followed by
verbatim frozen explanations of the top-five features.
Allocation and activation matrices appear in
Figure~\ref{fig:v2-case-diagnostics}.
$\dagger$ marks a feature outside $\mathcal S_{\mathrm{strong}}$.}
\label{tab:v2-case-details}

\begin{tabular}{@{}p{0.06\textwidth}p{0.13\textwidth}
p{0.75\textwidth}@{}}
\toprule

\multicolumn{3}{@{}l}{
\textbf{High revenue: \texttt{2x5c\_034928}}} \\
\midrule

\multicolumn{3}{@{}p{0.96\textwidth}@{}}{
\textit{Valuations, ordered by item 0--4:}
$\mathbf v_0=(0.488624,0.792755,0.0499006,0.370734,0.120901)$;
$\mathbf v_1=(0.781921,0.236717,0.640505,0.773145,0.567806)$.
} \\

\multicolumn{3}{@{}p{0.96\textwidth}@{}}{
\textit{Bidder accounting, ordered as $[b_0,b_1]$:}
$V=[0.817325,2.63371]$;
$p=[0.594104,2.03486]$;
$U=[0.223221,0.598849]$;
$P=[0.726888,0.772622]$;
$r=[5.35548\times10^{-4},0]$.
} \\

\midrule
Feature & Frozen target & Frozen explanation \\
\midrule

F509 & Utility $\uparrow$
& Allocating valuable items to the focal bidder while leaving
meaningful surplus raises truthful utility; this high-surplus
regime predicts activation of the feature. \\

F428 & Payment share $\downarrow$
& The feature activates when the focal bidder pays a relatively
small share of allocated value, indicating favorable surplus
from a low effective payment burden. \\

F419 & Payment share $\downarrow$
& The feature activates when the focal bidder pays only a small
share of substantial allocated value, and is usually absent
when payment consumes much of that value. \\

F335 & Utility $\uparrow$
& The feature activates when the focal bidder is a near-monopoly
winner whose dominant allocation and small payment produce
high truthful utility, unlike diffuse or costly allocations. \\

F132 & Margin $\uparrow$
& The feature is most likely active in decisive, near-exclusive
allocation regimes: large winner margins concentrate items
with one bidder, whereas diffuse or contested allocations
suppress activation regardless of whether the focal bidder wins. \\

\midrule
\multicolumn{3}{@{}l}{
\textbf{Near no trade: \texttt{2x5c\_010715}}} \\
\midrule

\multicolumn{3}{@{}p{0.96\textwidth}@{}}{
\textit{Valuations, ordered by item 0--4:}
$\mathbf v_0=(0.0633496,0.442359,0.0699195,0.0397434,0.487681)$;
$\mathbf v_1=(0.472124,0.0825441,0.128804,0.0445592,0.341537)$.
} \\

\multicolumn{3}{@{}p{0.96\textwidth}@{}}{
\textit{Bidder accounting, ordered as $[b_0,b_1]$:}
$V=[2.30854\times10^{-4},9.41062\times10^{-6}]$;
$p=[2.29665\times10^{-4},9.33243\times10^{-6}]$;
$U=[1.18924\times10^{-6},7.81929\times10^{-8}]$;
$P=[0.994849,0.991691]$;
$r=[0,0]$.
} \\

\midrule
Feature & Frozen target & Frozen explanation \\
\midrule

F283$\dagger$ & Payment share $\uparrow$
& The feature activates when the focal bidder-item lies in a
near-degenerate, high-payment-share regime, often with
effectively no allocation, rather than in ordinary competitive
allocations where payment is a smaller fraction of value. \\

F16 & Payment share $\uparrow$
& A near-total payment share compresses the focal bidder's
truthful surplus, and the feature marks these bidder-item
tokens; it is absent when payment consumes a smaller share. \\

F173 & Welfare $\downarrow$
& Activates in inefficient allocation regimes where probability
mass reaches lower-valued bidders, reducing truthful social
welfare; allocations aligned with high valuations more often
leave the feature inactive. \\

F76 & Revenue $\downarrow$
& The feature activates in a no-trade regime: the mechanism
allocates nothing and collects essentially no seller payment,
causing bidder-item tokens to share a global
inactive-auction representation. \\

F506 & Payment share $\uparrow$
& Feature 506 activates for focal bidder-item tokens in
payment-heavy allocations: the bidder pays a large share of
allocated value, compressing truthful surplus and
distinguishing these tokens from cheap decisive wins. \\

\bottomrule
\end{tabular}

\par\smallskip
\begin{minipage}{0.96\textwidth}
\scriptsize
$V$, $p$, $U$, $P=p/V$, and $r$ denote bidder-level allocated
value, payment, utility, payment share, and searched deviation gain,
respectively. Empirical regret averages the two deviation gains;
zero means no positive gain was found.
Feature rows follow the heatmap order, and arrows denote preassigned
association-derived directions.
\end{minipage}
\end{table}

\section{Cross-Setting Applicability}
\label{app:v2-additional-settings}
We report the cross-setting causal-response and empirical-regret audits referenced in Section~\ref{subsec:v2-usefulness}. Alongside the five-item continuous setting $2\times5$c, we evaluate the single-item continuous setting $2\times1$c and the five-item discrete-type setting $2\times5$d, which draws valuations from type-pair-specific truncated-normal distributions across ten bidder and ten item types. Each setting uses an independently trained mechanism, SAE dictionary, and feature explanations under the four-dose protocol. As shown in Table~\ref{tab:v2-setting-audits}, the strong-response criteria identify 165, 106, and 142 features in $2\times5$c, $2\times1$c, and $2\times5$d, covering 32.2\%, 20.7\%, and 27.7\% of each dictionary. Reoptimized misreport audits further identify both regret-increasing and regret-reducing features in every setting, with overall BH-significant counts of 26/17, 37/43, and 43/10 and payment-share counts of 17/11, 23/27, and 19/3 across $2\times5$c, $2\times1$c, and $2\times5$d. These results show that our pipeline consistently identifies effective setting-specific economic features and supports bidirectional incentive audits across the tested auction settings, demonstrating its robustness.

\end{document}